\documentclass[10pt,twocolumn,letterpaper]{article}

\usepackage[pagenumbers]{cvpr} 

\usepackage{graphicx}
\usepackage{amsmath}
\usepackage{amssymb}
\usepackage{booktabs}

\usepackage[pagebackref,breaklinks,colorlinks]{hyperref}

\usepackage[capitalize]{cleveref}
\crefname{section}{Sec.}{Secs.}
\Crefname{section}{Section}{Sections}
\Crefname{table}{Table}{Tables}
\crefname{table}{Tab.}{Tabs.}

\def\cvprPaperID{*****} 
\def\confName{CVPR}
\def\confYear{2022}

\begin{document}

\title{A Joint Cross-Attention Model for Audio-Visual Fusion in Dimensional \\Emotion Recognition}

\author{R~Gnana~Praveen,Wheidima~Carneiro~de~Melo, Nasib Ullah, Haseeb Aslam, Osama Zeeshan\\
Théo Denorme, Marco Pedersoli, Alessandro Koerich, Simon Bacon, Patrick Cardinal, and Eric Granger\\
Laboratoire d'imagerie, de vision et d'intelligence artificielle (LIVIA)\\
École de technologie supérieure, Montreal, Canada\\
{\tt\small gnanapraveen.rajasekar.1@ens.etsmtl.ca, wheidima.melo@oulu.fi, eric.granger@etsmtl.ca}
}
\maketitle

\begin{abstract}
Multimodal emotion recognition has recently gained much attention since it can leverage diverse and complementary relationships over multiple modalities, such as audio, visual, and biosignals. Most state-of-the-art methods for audio-visual (A-V) fusion  rely on recurrent networks or conventional attention mechanisms that do not effectively leverage the complementary nature of A-V modalities. In this paper, we focus on dimensional emotion recognition based on the fusion of facial and vocal modalities extracted from videos. Specifically, we propose a joint cross-attention model that relies on the complementary relationships to extract the salient features across A-V modalities, allowing for accurate prediction of continuous values of valence and arousal. The proposed fusion model efficiently leverages the inter-modal relationships, while reducing the heterogeneity between features. In particular, it computes cross-attention weights based on the correlation between joint feature representations, and that of individual modalities. By deploying a joint A-V feature representation into the cross-attention module, the performance of our fusion module improves significantly over the vanilla cross-attention module. Experimental results\footnote{The code is available on GitHub: \url{https://anonymous.4open.science/r/JointCrossAttentional-AV-Fusion-06F7}.} on the AffWild2 dataset highlight the robustness of our proposed A-V fusion model.  It has achieved a concordance correlation coefficient (CCC) of 0.374 (0.663) and 0.363 (0.584) for valence and arousal, respectively, on test set (validation set). This is a significant improvement over the baseline of third challenge of Affective Behavior Analysis in-the-wild (ABAW3) competition, with a CCC of 0.180 (0.310) and 0.170 (0.170). 
\end{abstract}

\section{Introduction}
\label{sec:intro}

\begin{figure}[!t]
\centering
\includegraphics[width=0.48\textwidth]{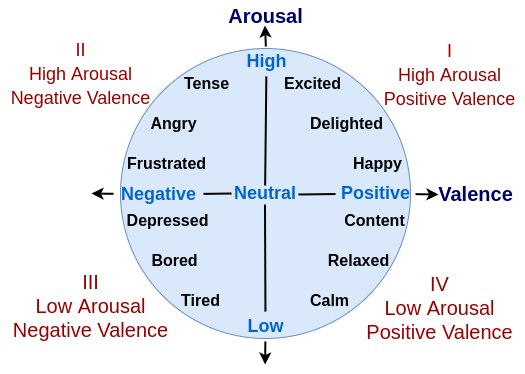}
\caption{\textbf{The valence-arousal space.}}
\label{fig:VA}
\end{figure}

Emotion recognition (ER) is a challenging problem since the expressions linked to human emotions are extremely diverse in nature across individuals and cultures. It has been extensively researched in various fields such as neuroscience, psychology, cognitive science and computer science, leading to the advancement of a wide range of applications in, e.g., health care (e.g., assessment of anger, fatigue, depression and pain), robotics (human-machine interaction), driver assistance (assessment of a driver’s state), etc \cite{Kolakowska2014}. ER can be formulated as the problem of categorical model or dimensional model of emotions. In categorical model, the human emotions has been categorized into six categories – anger, disgust, fear, happy, sad, and surprise \cite{Ekman}. Subsequently, contempt has been added to these six basic emotions \cite{Matsumoto}. The categorical model of ER has been explored extensively in the field of affective computing due to its simplicity and universality. In dimensional model, human emotions can be analyzed on a wide range of emotions on a continuous scale, where the human emotions can be projected onto the dimensions of valence and arousal \cite{Schlosberg}. Figure \ref{fig:VA} illustrates the use of a two-dimensional space to represent emotional states, where valence and arousal are employed as dimensional axes.   Valence reflects the wide range of emotions in the dimension of pleasantness from being negative (sad) to positive (happy), whereas arousal spans the range of intensities from passive (sleepiness) to active (high excitement).

Dimensional modelling of emotions is more challenging than categorical case since it is difficult to obtain continuous scale of annotations compared to discrete emotions. Due to the continuous range of emotions, the annotations tend to be noisy and ambiguous. Several databases such as RECOLA \cite{6553805}, SEWA \cite{8854185}, SEMAINE \cite{5959155}, etc have been introduced for the task of dimensional ER. Depending on the video capture conditions, i.e., whether controlled or in-the-wild environments, this task can present different challenges due to factors such as poor illumination, pose variations, and background noise. Recently, Kollias et al. \cite{Kollias} introduced Affwild2 database, which is the largest in-the-wild  database for dimensional ER. The dataset is also provided with the annotations of other tasks of expression classification and action unit detection. Previously, the data-set has been used for challenges hosted in conjunction with CVPR 2017 \cite{zafeiriou2017aff}, FG 2020 \cite{kollias2020analysing} and ICCV 2021 \cite{Kollias_2021_ICCV}. Several approaches have been proposed for previous challenges in the framework of multi-task learning  \cite{kollias2021distribution, kollias2021affect, kollias2019expression, kollias2019face}. In continuation with the previous challenges, third competition was held in conjunction with CVPR 2022 \cite{kollias2022abaw} with an exclusive challenge track for valence and arousal estimation. 

In this paper, we investigate the prospect of leveraging the complementary relationship of A and V modalities in videos in a joint cross attentional framework. Facial expressions are one of the most dominant channels through which human emotions can be expressed. It was shown that only one-third of human communication is conveyed through verbal components and two-third of communication occur through non-verbal components \cite{Mehrabian}. Voice also serves as a major cue in conveying human emotions as it often carry complementary relationship with the V modality. For instance, when the facial modality is missing due to pose, blur, low illumination, etc., we can still leverage the A modality to estimate the emotional state. Similarly, when we have silent regions in the A modality, we can leverage the rich information in the V modality. In most of the existing approaches, A-V fusion is often achieved by concatenating the A and V features, which may degrade system performance \cite{cite7}. Therefore, designing a fusion mechanism based on A and V features which can effectively leverage their complementary relationships is pivotal in improving the performance of a multimodal ER system over uni-modal approaches. 

Several ER approaches have been proposed for video-based dimensional ER using convolutional neural networks (CNNs) to obtain the deep features, and recurrent neural networks (RNNs) to capture the temporal dynamics \cite{cite6,cite7}. Deep models have also been widely explored for vocal emotion recognition, typically using spectrograms with 2D-CNNs \cite{cite6, 9607711}, or raw wave forms with 1D-CNNs \cite{cite7}. In most of the existing approaches \cite{cite7, cite8} for dimensional ER, A-V fusion is performed by concatenating the deep features extracted from individual facial and vocal modalities, and fed to LSTM for predicting valence and arousal. Although LSTM based fusion models the spatio-temporal and intra-modal relationships, and can improve system performance, it does not effectively capture the inter-modal relationships across the individual modalities. We therefore investigate the prospect of extracting more contributive features across A and V modalities in order to leverage their complementary temporal relationships. 

Attention mechanisms have recently gained much interest in the computer vision and machine learning communities, allowing to extract task relevant features, and thereby improving system performance. Most of the existing attention based approaches for dimensional ER explore the intra-modal relationships \cite{Jiyoung}. Although a few approaches  attempt to capture the cross-modal relationships using cross-attention based on transformers \cite{cite8, srini_2021_SLT}, they do not effectively leverage the complementary relationship of A-V modalities. Indeed, their computation of attention weights does not consider the correlation among the A and V features. Recently, Praveen et al. \cite{9667055} proposed cross-attentional model for dimensional ER based on AV fusion and showed significant improvement on RECOLA dataset \cite{6553805} over state-of-the-art methods by leveraging the complementary relationships of A and V modalities. In this paper, we introduce a joint modeling of intra- and inter-modal relationships into a cross attentional framework. The cross correlation is computed between the joint A-V feature representation, and the features of individual modalities. We have shown that deploying joint representation into the cross attentional module significantly improves the modeling of cross-modal relationships over the vanilla cross attentional model \cite{9667055}, while reducing the heterogeneity across the modalities on the challenging in-the-wild Affwild2 dataset\cite{Kollias}.  

The main contributions of the paper are as follows.
    (1) A joint cross-attentional model is proposed for A-V fusion based on the joint modeling of intra- and inter-modal relationships, which effectively captures the complementary inter-modal as well as intra-modal relationships. Specifically, we use joint A-V feature representations to attend to the other modality (as well as itself) based on the attention weights computed from the cross correlation between the individual features and joint representation.
    (2) The effectiveness of the proposed approach is analyzed through an extensive set of experiments and ablation studies on the challenging in-the-wild Affwild2 dataset. 


The rest of this paper is organized as follows. Section \ref{Related Work} provides a critical analysis of the relevant literature on dimensional ER, and attention models for A-V fusion. Section \ref{Proposed Approach} describes the proposed joint cross-attentional A-V fusion model. Sections \ref{Training details} and \ref{results} present the experimental methodology for validation, and results obtained with the proposed approach.  

\section{Related Work}
\label{Related Work}

\subsection{A-V Fusion Based Emotion Recognition}
One of the primitive approaches using DL models for A-V fusion based dimensional ER was proposed by Tzirakis et al. \cite{cite7}, where A and V features, obtained from ResNet50 and 1D-CNN respectively, are concatenated and fed to Long short-term memory model (LSTM). Juan et al. \cite{8914655} presented an empirical study of fine-tuning various layers of pretrained CNN models for V modality, and used conventional A features for fusion. Nguyen et al. \cite{9374787} proposed a deep model of two-stream auto-encoders and LSTM to simultaneously learn compact representative features from A and V modalities for dimensional ER. Schonevald et al. \cite{cite6} explored knowledge distillation using teacher-student model for V modality and CNN model for A modality using spectrograms, and fused using RNNs. Deng et al \cite{9607738} proposed iterative self distillation method for modeling the uncertainties in the labels in a multi-task framework. Kuhnke et al. \cite{9320301} proposed two stream A-V network, where V features are extracted from R(2plus1)D model pretrained from action recognition dataset and A features are obtained from Resnet18 model. Wang et al \cite{9607711} further improved their approach \cite{9320301} by introducing teacher-student model in a semi-supervised learning framework. The teacher model is trained on the available labels, which is further used to obtain pseudo labels for unlabeled data. The pseudo labels are finally used to train the student model, which is used for final prediction. 
Though the above mentioned approaches have shown significant improvement for dimensional ER, they fail to effectively capture the inter-modal relationships and relevant salient features specific to the task. Therefore, we have focused on capturing the comprehensive features in a complimentary fashion using attention mechanisms.  

\subsection{Attention Models for A-V Fusion:}
Attention models for A-V fusion has been widely explored in modeling the intra and inter modal relationships between A-V modalities for various applications such as A-V event localization \cite{9423042}, action localization \cite{lee2021crossattentional}, emotion recognition \cite{srini_2021_SLT}, etc. Zhang et al. \cite{9320215} proposed attentive fusion mechanism, where multi features are obtained from 3D-CNN and 2D-CNN for V modality and 2D-CNN using spectrograms for A modality. The obtained A and V features are further re-weighted using scoring functions based on the relevant information in the individual modalities. Recently, cross-modal attention is found to be promising as effective modeling of inter-modal relationships significantly improves the system performance. Srinivas et al. \cite{srini_2021_SLT} explored transformers with encoder layers, where cross-modal attention is deployed to integrate A and V features for dimensional ER. Tzirakis et al. \cite{cite8} investigated self attention as well as cross-attention fusion based on transformers in order to enable the extracted features of different modalities to attend to each other. Although these approaches have explored cross-modal attention with transformers, they fail to leverage semantic relevance among the A-V features based on cross-correlation. Zhang et al. \cite{9607460} investigated the prospect of improving the fusion performance over individual modalities and proposed leader-follower attentive fusion for dimensional ER. The obtained features are encoded and attention weights are obtained by combining the encoded A and V features. The attention weights are further attended on the V features and concatenated to the original V features for final prediction. 

Unlike prior approaches, we advocate for a simple yet efficient joint cross-attentional model based on joint modeling of intra and inter modal relationships between A and V modalities. Cross-attention has been successfully applied in several applications, such as weakly-supervised action localization \cite{lee2021crossattentional}, few-shot classification \cite{NEURIPS2019_01894d6f} and dimensional ER \cite{9320216}. In most of these cases, cross-attention has been applied across the individual modalities. Praveen et al. \cite{9667055} have shown significant improvement using cross attention based on cross correlation across the individual features. However, we have explored joint attention between individual and combined AV-features. By deploying the joint AV feature representation, we can effectively capture the intra and inter-modal relationships simultaneously by allowing interactions across the modalities as well as oneself. Recently, joint co-attention has also been explored by Duan et al. \cite{9423042} in a recursive fashion for A-V event localization and found to be promising in obtaining robust multimodal feature representations. 
In this paper, joint (combined) A-V features are extracted through cross-attention, where the features of each modality attend to themselves, as well as those of the other modality, through cross-correlation of the concatenated  A-V features, and features of individual modalities. By effectively leveraging the joint modeling of intra- and inter-modal relationships, the proposed approach can significantly improve system performance.

\section{Proposed Approach}
\label{Proposed Approach}

\subsection{Visual Network:}
Facial expressions in videos carry rich information pertinent to both appearance and temporal dynamics, which plays a crucial role in understanding the emotions of a person. Therefore, these spatial and temporal cues must be efficiently modeled in order to obtain robust feature representations suitable for ER. In the recent years, deep learning models have been widely explored for analyzing facial expressions in videos. In most of these approaches \cite{5740839, WOLLMER2013153}, 2D-CNN has been used in conjunction with Recurrent Neural Networks (RNN) to capture the spatial and temporal dynamics respectively. 3D-CNNs have also been widely explored especially for action recognition and found to be promising in simultaneously capturing the spatial and temporal dynamics. Inspired by the performance of 3D-CNNs, \cite{8578773} explored R(2plus1)D network pretrained on the Kinetics-400 action recognition dataset \cite{9607711, 9320301} and outperformed  conventional 2D-CNNs for dimensional ER on Affwild2 dataset. Recently, I3D have shown significant improvement for action recognition with fewer number of parameters than that of conventional 3D-CNNs while able to leverage the pretrained weights of 2D-CNN models. 
However, it fails to capture the long-term temporal dependencies. Temporal Convolutional Networks (TCN) were found to be efficient in capturing the long term temporal dependencies \cite{9607460}. Therefore, we have explored TCN in conjunction with I3D in order to leverage both long- and short-term temporal dynamics. We have also explored other visual backbones, such as R(2plus1)D network pretrained on the Kinetics-400 action recognition dataset \cite{9607711, 9320301}, and ResNet CNNs with GRU to obtain the V features and validate our fusion model (see implementation details in Section \ref{Training details}).

\subsection{Audio Network:}
Several low-level descriptors such as prosodic, excitation, MFCC and spectral descriptors have commonly been used as feature representations for A modality in ER \cite{Helang, 8914655}. With the advent of deep models, the performance of speech ER have been significantly improved using 1D-CNNs on raw A signals \cite{cite7} or 2D-CNN models on spectrograms \cite{cite6, 9607711} over the past few years. Compared to 1D-CNNs, 2D-CNNs using spectrograms have been widely explored in the literature of speech ER, as it was found to carry significant para-lingual information pertaining to the affective state of a person \cite{Ma2018}. Various 2D-CNN architectures such as VGGish \cite{9607460} and Resnet18 \cite{7780459} have been used to obtain robust feature representations of A modality for ER. Given the ubiquitous usage of spectrograms for extracting effective feature representations pertinent to affective state of a person, we have also used spectrograms with 2D-CNNs in our framework to validate the proposed fusion model (see implementation details in Section  \ref{Training details}).

\begin{figure*}[t!]
\centering
\includegraphics[width=1.1\linewidth]{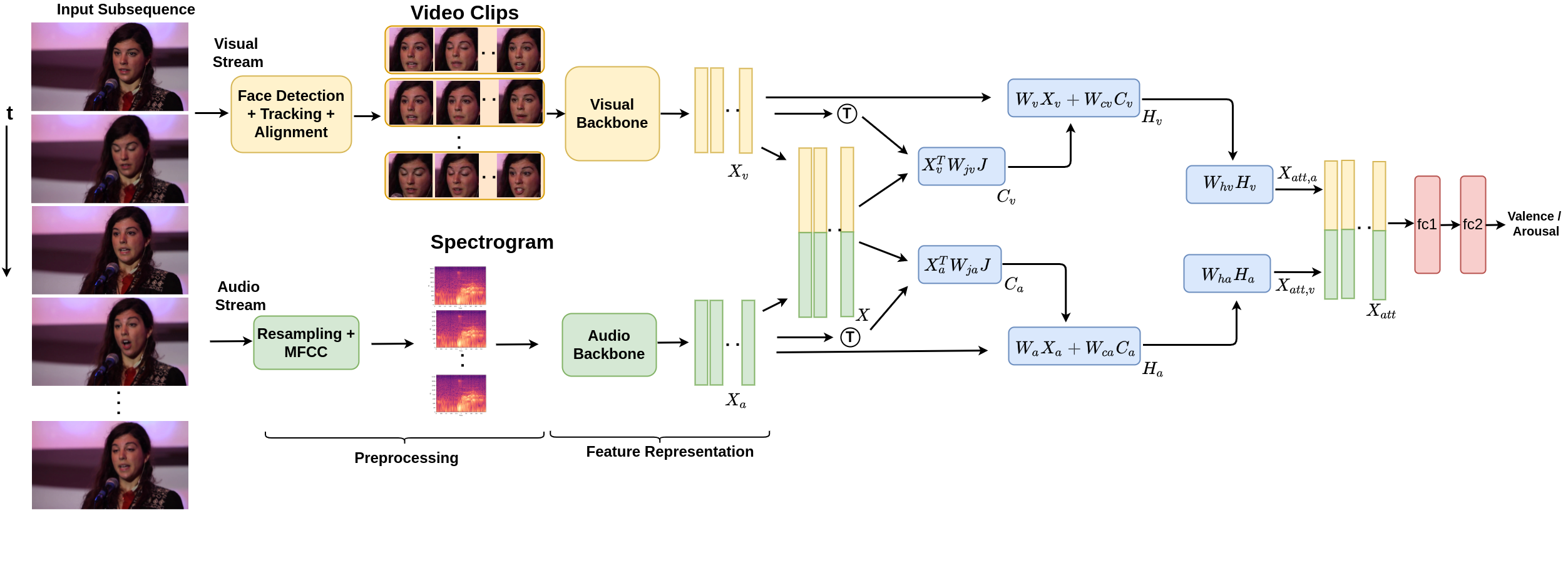}
\caption{\textbf{Joint cross-attention model proposed for A-V fusion (training mode).}}
\label{Block Diagram}
\end{figure*}

\subsection{Joint Cross-Attentional AV-Fusion:}

Though A-V fusion can be achieved through unified multimodal training, it was found that simultaneous training of multimodal networks often declines over that of individual modalities \cite{9156420}. This can be attributed to a number of factors, such as differences in learning dynamics for A and V modalities \cite{9156420}, different noise topologies, with some modality streams containing more or less information for the task at hand, as well as specialised input representations \cite{Nagrani21c}. Therefore, we have trained DL models for the individual A and V modalities independently in order to extract A and V features, which is fed to the joint cross-attentional module for A-V fusion that outputs final valence and arousal prediction.

For a given video sequence, the V modality carries more relevant information in some video clips, whereas A modality might be more relevant for others. Since multiple modalities convey diverse information for valence and arousal, their complementary relationship needs to be effectively captured. In order to reliably fuse these modalities, we rely on cross-attention based fusion mechanism to efficiently encode the inter-modal information, while preserving the intra-modal characteristics. Though cross-attention has been conventionally applied across the features of individual modalities, we used cross-attention in a joint learning framework. Specifically, our joint A-V feature representation is obtained by concatenating the A and V features to attend to the individual A and V features. By using the joint representation, features of each modality attend to them-self, and the other modality, helping to capture the semantic inter-modal relationships across A and V. The heterogeneity among the A and V modalities can also be drastically reduced by using the combined feature representation in the cross-attentional module, which further improves system performance. A block diagram of the proposed model is shown in Figure \ref{Block Diagram}.  

\noindent \textbf{A) Training mode:}  
Let ${\boldsymbol X}_{\mathbf a}$ and ${\boldsymbol X}_{\mathbf v}$ represents two sets of deep feature vectors extracted for the A and V modalities, in response to a given input video sub-sequence $\boldsymbol S$ of fixed size, where 
${\boldsymbol X}_{\mathbf a}\boldsymbol =  \{ \boldsymbol x_{\mathbf a}^1, \boldsymbol x_{\mathbf a}^2, ..., \boldsymbol x_{\mathbf a}^L \boldsymbol \} \in \mathbb{R}^{d_a\times L}$ and
${\boldsymbol X}_{\mathbf v}\boldsymbol =  \{ \boldsymbol x_{\mathbf v}^1, \boldsymbol x_{\mathbf v}^2, ..., \boldsymbol x_{\mathbf v}^L \boldsymbol \} \in \mathbb{R}^{d_v\times L}$. $L$ denotes the number of non overlapping fixed-size clips sampled uniformly from $\boldsymbol S$, ${d_a}$ and ${d_v}$ represents the feature dimension of the A and V representations, $\boldsymbol x_{\mathbf a}^{ l}$ and $\boldsymbol x_{\mathbf v}^{ l}$ denotes the A and V feature vectors, respectively, for $l = 1, 2, ..., L$ clips. 

As shown in Figure \ref{Block Diagram}, the joint representation of A-V features, $\boldsymbol{J}$, is obtained by concatenating the A and V feature vectors:  
 ${\boldsymbol J} = [{\boldsymbol X}_{\mathbf a} ; {\boldsymbol X}_{\mathbf v}] \in\mathbb{R}^{d\times L}$, 
where $d = {d_a} + {d_v}$ denotes the feature dimension of concatenated features. This A-V feature representations ($\boldsymbol J$) of the given video sub-sequence ($\boldsymbol S$) is now used to attend to unimodal feature representations ${\boldsymbol X}_{\mathbf a}$ and ${\boldsymbol X}_{\mathbf v}$. The joint correlation matrix $\boldsymbol C_{\mathbf a}$ across the A features ${\boldsymbol X}_{\mathbf a}$, and the combined A-V features $\boldsymbol J$ are given by: 
\begin{equation}
   \boldsymbol C_{\mathbf a}= \tanh \left(\frac{{\boldsymbol X}_{\mathbf a}^T{\boldsymbol W}_{\mathbf j \mathbf a}{\boldsymbol J}}{\sqrt d}\right)
\end{equation}
where ${\boldsymbol W}_{\mathbf j \mathbf a} \in\mathbb{R}^{L\times L} $ represents learnable weight matrix across the A and joint A-V features, and $T$ denotes transpose operation. Similarly, the joint correlation matrix for V features are given by: 
\begin{equation}
   \boldsymbol C_{\mathbf v}= \tanh \left(\frac{{\boldsymbol X}_{\mathbf v}^T{\boldsymbol W}_{\mathbf j \mathbf v}{\boldsymbol J}}{\sqrt d}\right)
\end{equation}

The joint correlation matrices $\boldsymbol C_{\mathbf a}$ and $\boldsymbol C_{\mathbf v}$ for A and V modalities provide  a semantic measure of relevance not only across the modalities but also within the same modality. Higher correlation coefficient of the joint correlation matrices $\boldsymbol C_{\mathbf a}$ and $\boldsymbol C_{\mathbf v}$ shows that the corresponding samples are strongly correlated within the same modality as well as other modality. 
Therefore, the proposed approach is able to efficiently leverage the complimentary nature of A and V modalities (i.e., inter-modal relationship) as well as intra-modal relationships, thereby improving the performance of the system. After computing the joint correlation matrices, the attention weights of A and V modalities are estimated. 

Since the dimensions of joint correlation matrices ($\mathbb{R}^{d_a\times d}$) and the features of corresponding modality ($\mathbb{R}^{L\times d_a}$) differ, we rely on a different learnable weight matrices corresponding to features of the individual modalities, in order to compute attention weights of the modalities. For the A modality, the joint correlation matrix $\boldsymbol C_{\mathbf a}$ and the corresponding A features ${\boldsymbol X}_{\mathbf a}$ are combined using the learnable weight matrices $\boldsymbol W_{\mathbf c \mathbf a}$ and $\boldsymbol W_{\mathbf a}$ respectively to compute the attention weights of A modality, which is given by 
\begin{equation}
\boldsymbol H_{\mathbf a}=ReLu(\boldsymbol W_{\mathbf a} \boldsymbol X_{\mathbf a}\;+\; \boldsymbol W_{\mathbf c \mathbf a} {\boldsymbol C}_{\mathbf a}^T)
\end{equation}
where ${\boldsymbol W}_{\mathbf c \mathbf a} \in\mathbb{R}^{k\times d} $, ${\boldsymbol W}_{\mathbf a} \in\mathbb{R}^{k\times L}$ and ${\boldsymbol H}_{\mathbf a}$ represents the attention maps of the A modality. Similarly, the attention maps ($\boldsymbol H_{\mathbf v}$) of V modality are obtained as 
\begin{equation}
\boldsymbol H_{\mathbf v}=ReLu(\boldsymbol W_{\mathbf v} \boldsymbol X_{\mathbf v}\;+\; \boldsymbol W_{\mathbf c \mathbf v} {\boldsymbol C}_{\mathbf v}^T)
\end{equation}
where ${\boldsymbol W}_{\mathbf c \mathbf v} \in\mathbb{R}^{k\times d} $, ${\boldsymbol W}_{\mathbf v} \in\mathbb{R}^{k\times L}$. 

Finally, the attention maps are used to compute the attended features of A and V modalities. These features are obtained as:  
\begin{equation}
{\boldsymbol X}_{\mathbf a \mathbf t \mathbf t, \mathbf a} = \boldsymbol W_{\mathbf h \mathbf a}  \boldsymbol H_{\mathbf a} + \boldsymbol X_{\mathbf a}
\end{equation}
\begin{equation}
{\boldsymbol X}_{\mathbf a \mathbf t \mathbf t, \mathbf v} = \boldsymbol W_{\mathbf h \mathbf v}  \boldsymbol H_{\mathbf v} + \boldsymbol X_{\mathbf v}  
\end{equation}
where $\boldsymbol W_{\mathbf h \mathbf a} \in\mathbb{R}^{k\times L}$ and $\boldsymbol W_{\mathbf h \mathbf v} \in\mathbb{R}^{k\times L}$ denote the learnable weight matrices, respectively. The attended A and V features, ${\boldsymbol X}_{\mathbf a \mathbf t \mathbf t, \mathbf a}$ and $ {\boldsymbol X}_{\mathbf a \mathbf t \mathbf t, \mathbf v}$ are further concatenated to obtain the A-V feature representation, which is given by:  
\begin{equation}
 {\boldsymbol X}_{\mathbf a\mathbf t\mathbf t} = [{\boldsymbol X}_{\mathbf a\mathbf t\mathbf t\boldsymbol,\mathbf v} ; {\boldsymbol X}_{\mathbf a\mathbf t\mathbf t\boldsymbol,\mathbf a} ]  
\end{equation}
Finally, the A-V features are fed to the fully connected layers for the predictions of valence or arousal.

The Concordance Correlation Coefficient ($\rho_c$) has been widely used in the literature to measure the level of agreement between the predictions ($x$) and ground truth ($y$) annotations for dimensional ER \cite{cite7}. Let $\mu_x$ and $\mu_y$ represents the mean of predictions and ground truth, respectively. Similarly, if $\sigma_x^2$ and $\sigma_y^2$ denotes the variance of predictions and ground truth, respectively, then $\rho_c$ between the predictions and ground truth is:
\begin{equation}
\rho_c=\frac{2\sigma_{xy}^2}{\sigma_x^2+\sigma_y^2+(\mu_x-\mu_y)^2}
\end{equation}
where $\sigma_{xy}^2$ denotes the covariance between predictions and ground truth. 
Although MSE has been widely used as a loss function for regression models, we use $\mathcal{L} = 1 - \rho_c$ since it is standard and common loss in the  dimensional ER literature \cite{cite7}. The parameters of our A-V fusion model ($\boldsymbol W_{\mathbf c \mathbf a}$, $\boldsymbol W_{\mathbf a}$, ${\boldsymbol W}_{\mathbf c \mathbf v}$, ${\boldsymbol W}_{\mathbf v}$, $\boldsymbol W_{\mathbf h \mathbf a}$, and $\boldsymbol W_{\mathbf h \mathbf v}$) are optimized according to this loss.

\noindent \textbf{B) Test mode:}  A continuous video sequence is input to our model during inference. Feature representations $\boldsymbol x_{\mathbf a}^l$  and $\boldsymbol x_{\mathbf v}^l$ are extracted by A and V backbones for successive input clips and spectrograms, and fed to the A-V fusion model for the prediction of valence and arousal.  In addition, the arousal and valence predictions may be produced using multiple diverse A and V backbones that are combined through feature-level fusion, or multiple A-V fusion models that are combined through decision-level fusion (see implementation details in Section \ref{Training details}).

\section{Experimental Methodology}
\label{Training details}

\subsection{Dataset:}
Affwild2 is the largest database in the field of affective computing captured from YouTube, under extreme challenging environments. Though the dataset is provided with annotations for the tasks of expression classification, action unit detection and valence-arousal, we have focused on the problem of estimating valence-arousal in this work. For the track of valence-arousal estimation challenge, there are $567$ videos with the annotations of valence and arousal. Sixteen of these video clips display two subjects, both of which have been annotated. The annotations are provided by four experts using a joystick and the final annotations are obtained as the average of the four raters. In total, there are $2, 786, 201$ frames with $455$ subjects, out of which $277$ are male and $178$ female. The annotations for valence and arousal are provided continuously in the range of [-1, 1]. Some of the frames in some videos are not annotated. So we discard those frames. The dataset is split into the training, validation and test sets. The partitioning is done in a subject independent manner, so that every subject’s data will present in only one subset. The partitioning produces $341$, $71$, and $152$ train, validation and test videos respectively.

\subsection{Implementation Details:}
For the \textbf{V modality}, we have used the cropped and aligned images provided by the challenge organizers \cite{Kollias_2021_ICCV}. For the missing frames in the V modality, we have considered black frames (i.e., zero pixels). Faces are resized to $224$x$224$ to be fed to the I3D network. The videos are converted to sub-sequences, which is further sampled uniformly to obtain non overlapping fixed-size clips. The subsequence length and the clip length of the videos are considered to be $64$ and $8$ respectively, obtained by down-sampling a sequence of $256$ frames by $4$. Therefore, we have 8 clips in each sub-sequence, resulting in $1,96,265$ training samples and $41,740$ validation samples and $92,941$ test samples. I3D model was pre-trained on ImageNet, and inflated to a 3D-CNN using Affwild2 videos of facial expressions. To regularize the network, dropout is used with $p = 0.8$ on the linear layers. The initial learning rate was set to be $1e-3$, and the momentum of $0.8$ is used for SGD. Weight decay of $5e-4$ is used. Here again, the batch size of the network is set to be $8$. Data augmentation is performed on the training data by random cropping, which produces scale invariant model. The number of epochs is set to 50, and early stopping is used to obtain weights of the best model.

For the \textbf{A modality}, the vocal signal is extracted from the corresponding video, and re-sampled to $44100$Hz, which is further processed to extract short vocal segments corresponding to a clip size of $32$ frames of the V network. It is ensured that the clips and sub-sequences of the V clips are properly synchronized with that of A clips. The spectrogram is obtained using Discrete Fourier Transform (DFT) of length 1024 for each short clip (corresponding to 32 frames), where the window length is considered to be 20 msec and the hop length to be 10 msec. Following aggregation of short-time spectra, we obtain the spectrogram of 64 x 107 corresponding to each sub-sequence of the V modality. The spectrogram is converted to log-power-spectrum, expressed in dB. Finally, mean and variance normalization is performed on the spectrogram. Now the obtained spectrograms are fed to the Resnet18 \cite{7780459} to obtain the A features. Due to the availability of the large number of samples in the Affwild2 dataset, we trained the Resnet18 model from scratch. In order to adapt to the number of channels of the spectrogram, the first convolutional layer in the Resnet18 model is replaced by single channel. The network is trained with an initial learning rate of $0.001$ and weights are optimized using Adam optimizer. The batch size is considered to be $64$ and early stopping is used to obtain the best model for prediction. 

For the \textbf{A-V fusion network}, the size of the concatenated A-V features $J$ are set to be $1024$. In the joint cross-attention module, the initial weights of the cross-attention matrix is initialized with Xavier method \cite{pmlr-v9-glorot10a}, and the weights are updated using Adam optimizer. The initial learning rate is set to be $0.001$ and batch size is fixed to be $64$. Also, dropout of $0.5$ is applied on the attended A-V features and weight decay of $5e-4$ is used for all the experiments. Feature-lever (decision-level) fusion is implemented by training a fully connected neural network to provide a weighted fusions of feature representations (decisions values) for arousal and valence predictions.
\section{Results and Discussion}
 \label{results}
\begin{table*}
    \centering
      \caption{\textbf{Performance of our approach model with various components on the Affwild2 dataset. Resnet18 \cite{7780459} is used to extract A features in all experiments.}}
    \label{Affwild2 visual network results}
    \begin{tabular}{|l|c|c|c|c||c|c|c|c|c|c|} 
	\hline
	 \textbf{V Backbone}  & \textbf{Fusion } &  \textbf{Valence} & \textbf{Arousal} \\
	\hline \hline
	I3D &  Feature Concatenation            & 0.531 &  0.468 \\
	\hline
	R3D &  Feature Concatenation            & 0.517 &  0.493 \\
	\hline
	I3D &  Cross-Attention \cite{9667055}   & 0.541 &  0.517 \\
    \hline
	I3D &  Leader-Follower \cite{cite6}     & 0.592 & 0.521  \\
	\hline \hline
	Resnet18 + GRU  &  Joint Cross Attention (Ours)           & 0.632  & 0.520 \\
    \hline 
	R3D &  Joint Cross-Attention (Ours)                      & 0.642 & \textbf{0.592}  \\
	\hline 
	I3D &  Joint Cross-Attention (Ours)                      & 0.657 & 0.580 \\
	\hline
	I3D + TCN &  Joint Cross-Attention (Ours)                & \textbf{0.663} &  0.584 \\
	\hline 

\end{tabular}
\end{table*}


\begin{table*}
\renewcommand{\arraystretch}{1.4}
    \centering
    \caption{ \textbf{CCC performance of the proposed and state-of-art methods for A-V fusion on the Affwild2 development set. }}
    \label{Comparison with state-of-the-art for Affwild2 validation}
    \begin{tabular}{|l||c|c|c|c|c|c||c|c|c|} 
	\hline
	 \textbf{Method -- A/V backbone}  &  \multicolumn{3}{|c|}{\textbf{Valence}} & \multicolumn{3}{|c|}{\textbf{Arousal}}  \\ \cline{2-7}
	 & \textbf{Audio}  & \textbf{Visual}  & \textbf{Fusion}  & \textbf{Audio} & \textbf{Visual} & \textbf{Fusion}\\
	 \hline
	\hline
    Kuhnke \cite{9320301}, FGW 2020 -- A: Resnet18; V: R(2plus1)D & 0.351 & 0.449 & 0.493 & 0.356 & 0.565 & 0.604 \\
	\hline
    Zhang \cite{9607460}, ICCVW 2021 -- A: VGGish; V: Resnet50 + TCN & - & 0.405 & 0.457 & - &  0.635  & \textbf{0.645} \\
    \hline
	 Rajasekhar \cite{9667055}, FG 2021 -- A: Resnet18; V: I3D + TCN & 0.351  & 0.417 & 0.552 & 0.356 & 0.539 & 0.531\\ 
	\hline 
	Joint Cross-Attention (Ours) -- A: Resnet18; V: I3D + TCN & 0.351  & 0.417 & \textbf{0.663} & 0.356 & 0.539 & 0.584\\ 
	\hline
\end{tabular}
\end{table*}

\begin{table*}
\renewcommand{\arraystretch}{1.4}
    \centering
    \caption{ \textbf{CCC performance of the proposed and state-of-art methods for A-V fusion on Affwild2 test set.}}
    \label{Comparison with state-of-the-art for Affwild2 test}
    \begin{tabular}{|l||c|c|c|c|c|c||c|c|c|} %
	\hline
	 \textbf{Method } & \textbf{Modalities Used} 
	 & \textbf{Valence}  &  \textbf{Arousal} & \textbf{Mean}\\
	 \hline  \hline
	Situ-RUCAIM3 \cite{Situ-RUCAIM3}  & Audio, Visual  & 0.606 & 0.596 & 0.601 \\
	\hline
    FlyingPigs \cite{FlyingPigs} & Audio, Visual, Text & 0.520 & 0.602 &  0.561\\
	\hline
	PRL \cite{PRL} & Visual & 0.450 & 0.445 & 0.448 \\
	\hline
	HSE-NN \cite{HSE-NN} & Visual & 0.417 & 0.454 &  0.436\\
	\hline
	AU-NO \cite{AU-NO} & Audio, Visual & 0.418 & 0.407 & 0.413 \\
	\hline
   Joint Cross-Attention (Ours) & Audio, Visual & \textbf{0.374} & \textbf{0.363} &  \textbf{0.369} \\
	\hline
    Baseline & Visual & 0.180 & 0.170 &  0.175 \\
	\hline
\end{tabular}
\end{table*}


\subsection{Ablation Study}
Table \ref{Affwild2 visual network results} presents the results of our ablation study on the validation dataset. The performance of our proposed joint cross-attentional fusion is compared using various A and V backbones and A-V fusion strategies. First, we have implemented I3D \cite{8099985} with simple feature concatenation, where the extracted A and V features are concatenated, and fed to fully connected layers for valence and arousal prediction. Then we have replaced I3D with R3D \cite{8578773} and implemented a similar fusion strategy of feature concatenation. R3D was found to perform slightly better than I3D for arousal while I3D shows superior performance for valence. We have also compared our proposed approach with that of other relevant attention fusion strategies in the literature. We have compared the backbones of I3D with that of leader-follower attention \cite{9607460} and cross-attention \cite{9667055}. When compared to vanilla cross attention model, leader-follower attention was found to perform better. 

Finally, in order to validate the generalization capability of the proposed fusion model we have implemented various V backbones of I3D, R3D, Resnet18  with GRU and I3D with TCN.  Though the performance of our fusion model slightly varies with different backbones, we can observe that the proposed fusion model is able to achieve superior performance over that of other attention strategies \cite{9607460, 9667055} especially for valence. Compared to 2D-CNN model (Resnet18 with GRU), 3D-CNNs are found to perform slightly better. I3D shows improvement over valence than arousal with our fusion model compared to R3D. By introducing TCN with I3D, the performance of the proposed fusion model is found to perform even better as it captures better long term temporal cues than vanilla I3D. For all the experiments conducted above, Resnet18 is used as the backbone for the A modality.

\subsection{Comparison to state-of-the-art}
Table \ref{Comparison with state-of-the-art for Affwild2 validation} shows our comparative results against relevant state-of-the-art A-V fusion models on the Affwild2 validation set submitted for the previous challenges \cite{kollias2020analysing,Kollias_2021_ICCV}. Most of the relevant approaches have been implemented with different experimental protocol and training strategies. Therefore, in order to have a fair comparision we have re-implemented these approaches according to our experimental protocol, and analyzed the results on Affwild2 validation set. Similar to our A and V backbones, Kuhnke et al \cite{9320301} also used 3D-CNNs, where R(2plus1)D model is used for V modality and Resnet18 is used for A modality. However, they use additional masks for V modality and annotations of other tasks to refine the annotations of valence and arousal. They further perform simple feature concatenation without any specialized fusion model for the prediction of valence and arousal. So the fusion performance was not significantly improved over the uni-modal performance. Zhang et al \cite{9607460} explored leader follower attention model for fusion and showed minimal improvement of fusion performance over uni-modal performances. Though they have shown significant performance for arousal than valence, it is highly attributed to the V backbone. In our proposed approach, we have shown significant improvement for fusion especially for valence than arousal. Even with vanilla cross attentional fusion \cite{9667055}, we have shown that fusion performance for valence has been improved better than \cite{9607460} and \cite{9607460}. By deploying joint representation into the cross attententional fusion model, the fusion performance of valence has been significantly improved further. In case of arousal, though the fusion performance is lower than that of \cite{9607460} and \cite{9607460}, we can observe that it has been improved better than that of uni-modal V performance. Therefore, the proposed approach is effective in capturing the variations spanning over a wide-range of emotions (valence) than that of the intensities of the emotions (arousal). 

We have further compared our fusion model with that of other valid submissions for the third ABAW challenge \cite{kollias2022abaw} on the test set as shown in Table \ref{Comparison with state-of-the-art for Affwild2 test}. The winner of the challenge \cite{Situ-RUCAIM3} also uses A-V fusion and showed outstanding performance for both valence and arousal. They have used three external datasets to improve the generalization capability of the training model and features from multiple backbones for both V and A modalities. FlyingPigs \cite{FlyingPigs} uses text modality along with A and V modalities and showed improvement over A-V fusion using leader follower attention strategy. Apart from these, AU-NO \cite{AU-NO} is the only approach that relies on A-V fusion. 
They have investigated the performance of attention mechanisms such as self attention and cross attention with that of recurrent networks. They have further used additional loss components of Mean Square Error (MSE) and categorical cross entropy loss along with CCC. PRL \cite{PRL} and HSE-NN\cite{HSE-NN} uses only visual modality, where \cite{PRL} uses ensemble based strategy and \cite{HSE-NN} uses external AffectNet dataset \cite{8013713} for better performance. It is worth mentioning that we don't use any advanced loss components or post processing operations on predictions using cross validation, etc. apart from clipping the predictions to the range of [-1,1]. We also do not use any external data-sets or features from multiple backbones for A and V modalities. The performance of the proposed approach is solely attributed to the efficacy of our fusion model. We can observe that the fusion performance has been significantly improved over the uni-modal performances especially for valence. The proposed fusion model can be further improved using fusion of multiple A and V backbones either through feature level or decision level fusion similar to that of the winner of the challenge \cite{Situ-RUCAIM3}.     

\section{Conclusion}
\label{Conclusion}
In this work, joint cross-attentional is introduced for A-V fusion in video-based dimensional ER, leveraging the intra- and inter-modal relationships across A and V features. In particular, the complimentary relationship between A and V features are efficiently captured based on the correlation between the combined A-V features and individual A and V features. By jointly modeling the inter and inter-modal relationships, features of each modality attend to the other modality as well as itself, resulting in robust A and V feature representations. With the proposed model, A and V backbones are first trained individually for facial (V) and vocal (A) modalities.  Then, an attention mechanism based on correlation between joint and individual features are applied to obtain the attended A and V features. Finally, the attention weighted features are concatenated, and fed to linear connected layers to predict valence and arousal values. The proposed A-V fusion model is validated experimentally on the challenging Affwild2 video datasets, using different A and V backbones. Results show that the proposed model is a cost-effective approach that can sustaining a high level of performance, and outperform the state-of-the-art.



{\small
\bibliographystyle{ieee_fullname}
\bibliography{ABAWChallenge}
}

\end{document}